\documentclass[10pt,twocolumn,letterpaper]{article}

\usepackage[pagenumbers]{cvpr} 

\usepackage[dvipsnames]{xcolor}

\usepackage[accsupp]{axessibility}
\usepackage{soul}
\usepackage{url}
\usepackage{algorithm}
\usepackage{algpseudocode}
\usepackage{graphicx}
\usepackage{booktabs}
\usepackage{array}
\usepackage{microtype}
\usepackage{graphicx}
\usepackage{xcolor}
\usepackage{amsmath}
\usepackage{bm}
\usepackage{dsfont}
\usepackage{mathtools}
\usepackage{nccmath}
\usepackage{amssymb}
\usepackage{multirow}
\usepackage{booktabs} 
\usepackage{varwidth}
\usepackage{pifont}
\usepackage{makecell}
\usepackage[font=small,labelfont=bf,tableposition=top]{caption}
\usepackage{wrapfig}
\usepackage[flushleft]{threeparttable}
\usepackage{varwidth}
\usepackage{pifont}
\usepackage{makecell}
\usepackage{enumitem}
\usepackage{soul}

\newcommand{\Wb}{\mathbf{W}}

\usepackage{soul}
\usepackage{makecell}

\definecolor{cvprblue}{rgb}{0.21,0.49,0.74}
\usepackage[pagebackref,breaklinks,colorlinks,citecolor=cvprblue]{hyperref}

\def\paperID{2756} 
\def\confName{CVPR}
\def\confYear{2024}

\title{Transferable and Principled Efficiency for Open-Vocabulary Segmentation}
\author{Jingxuan Xu$^1$, Wuyang Chen$^2$, Yao Zhao$^{1,3}$, Yunchao Wei$^{1,3}$\\
$^1$Beijing Jiaotong University\quad $^2$Simon Fraser University\quad $^3$Peng Cheng Laboratory\\
}

\begin{document}
\maketitle
\begin{abstract}
Recent success of pre-trained foundation vision-language models makes Open-Vocabulary Segmentation (OVS) possible.
Despite the promising performance, this approach introduces heavy computational overheads for two challenges: 1) {large model sizes} of the backbone; 2) expensive costs during the {fine-tuning}.
These challenges hinder this OVS strategy from being widely applicable and affordable in real-world scenarios.
{
Although traditional methods such as model compression and efficient fine-tuning can address these challenges, they often rely on heuristics. This means that their solutions cannot be easily transferred and necessitate re-training on different models, which comes at a cost.
}
In the context of efficient OVS, we target achieving performance that is comparable to or even better than prior OVS works based on large vision-language foundation models, by utilizing \textbf{smaller models that incur lower training costs}. 
{
The core strategy is to make our efficiency \textbf{principled} and thus seamlessly \textbf{transferable} from one OVS framework to others without further customization.
}
Comprehensive experiments on diverse OVS benchmarks demonstrate our superior trade-off between segmentation accuracy and computation costs over previous works.
Our code is available on \url{https://github.com/Xujxyang/OpenTrans}
\end{abstract}
    
\begin{figure}[h]
\centering
\begin{center}
   \includegraphics[width=1.0\linewidth]{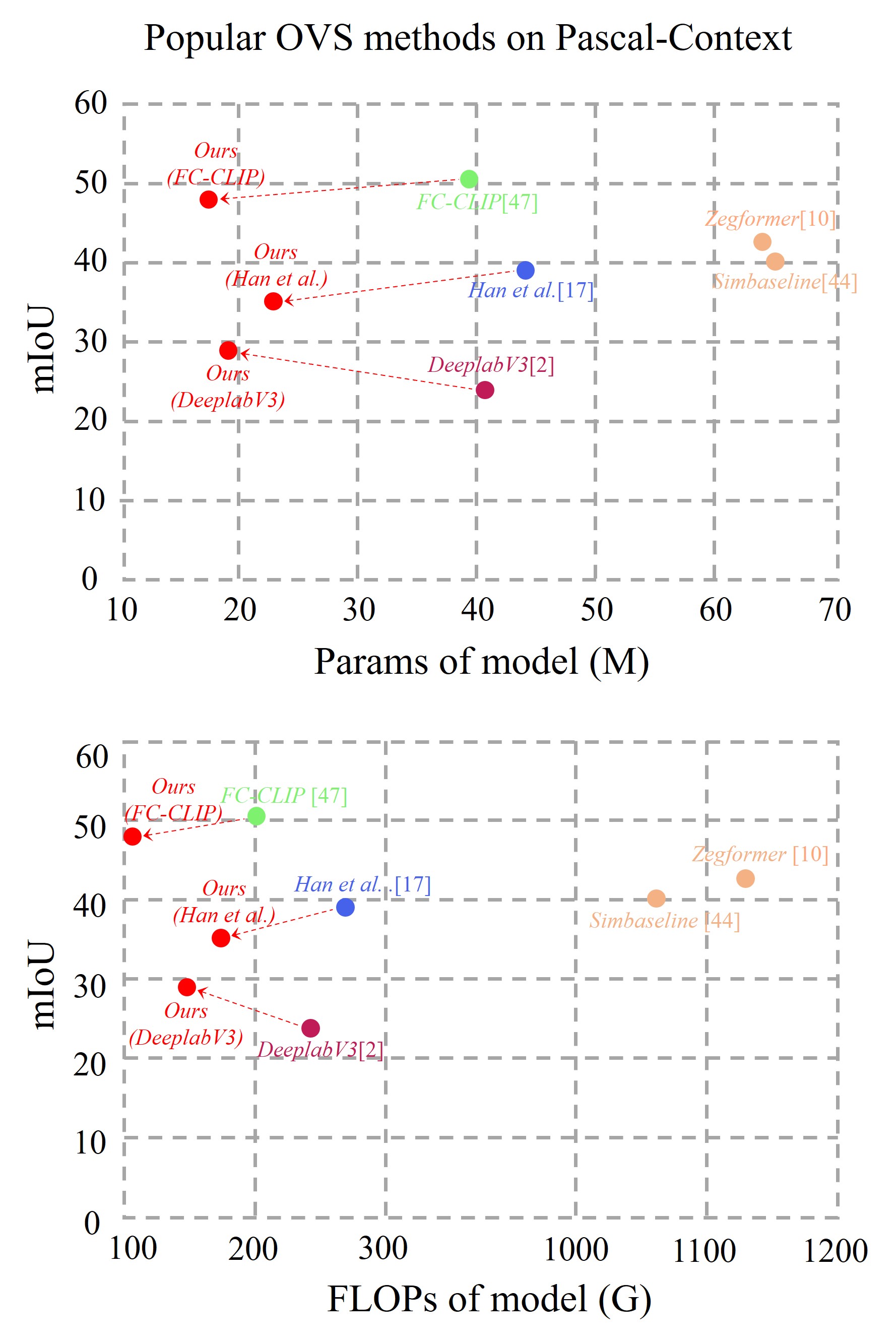}  
\end{center}
\vspace{-6mm}
\caption{Comparison with popular open-vocabulary segmentation works on Pascal-Context~\cite{mottaghi2014role} (with Resnet50 as backbone). The \textcolor{red}{red} points represent our works, the \textcolor{orange}{orange} points stand for traditional two-stage open-vocabulary segmentation~\cite{ding2022decoupling, xu2022simple}, the \textcolor{blue}{blue} point stands for single-stage open-vocabulary segmentation ~\cite{han2023global}, the \textcolor{green}{green} denotes a novel dual-channel prediction model with frozen backbone~\cite{yu2024convolutions}, the \textcolor{purple}{purple} represent traditional convolutional segmentation networks~\cite{chen2017rethinking}.}
\vspace{-6mm}
\label{fig:slot}   
\end{figure} 

\section{Introduction}

Open-vocabulary segmentation has been developed to address the limitations of traditional semantic segmentation~\cite{chen2017rethinking, wei2018revisiting, huang2019ccnet, zhao2017pyramid}, which cannot identify categories beyond its training set, by leveraging human-like recognition of novel categories using arbitrary text descriptions. This method enables the segmentation of arbitrary categories from text inputs, expanding its use in areas like image editing and human-robot interaction. Early models used cross-modal alignment between pixel and text embeddings~\cite{xian2019semantic,pmosr,li2022languagedriven}, while recent methods have improved performance with region-level alignment~\cite{ding2022decoupling,openseg,liang2023open,D2Zero,xu2022simple}. These methods learn from base class embeddings and are often paired with a frozen pre-trained vision-language foundation model, such as CLIP~\cite{radford2021learning}, to assist in re-classification, with strategies that incorporate knowledge distillation or a frozen CLIP backbone to maintain a generalizable representation while fine-tuning on known classes.


Despite the promising performance of recent OVS frameworks, these methods introduce heavy computational overheads.
The core bottleneck to the efficiency of recent OVS frameworks is rooted in the challenging balance between the preservation of the generalization ability of pretrained CLIP model versus the overhead of model size and training costs.
\textbf{First}, as the pretrained CLIP model supports the extraction of generalizable features of unseen images, the entire heavy image encoder has to be inherited for the OVS task. This introduces the requirement for the \ul{large model size} of the CLIP image encoder.
\textbf{Second}, after inheriting from CLIP, one needs to fine-tune  together with the segmentation head. This further requires \ul{expensive training computations}, and careful and delicate fine-tuning to preserve the pretrained generalizable knowledge in the CLIP image encoder.
These challenges hinder this OVS strategy from being more widely applicable and affordable in real-world scenarios.

Traditional methods, including model compression~\cite{frankle2018lottery,chen2020lottery,chen2021lottery} and efficient fine-tuning~\cite{rusu2016progressive,houlsby2019parameter,hu2021lora}, can mitigate these challenges. For example, iterative magnitude pruning can find semantic-aware lottery tickets that introduce highly sparse networks adapted to downstream tasks. Adapter-based fine-tuning can partially update pretrained models by introducing lightweight modules.
Despite the efficiency, however, these methods come with a price of \textbf{heuristics}: their solutions cannot be easily transferred and have to be re-trained on different models.
One has to re-run the pruning to find task-specific and model-specific sparse masks that can \textit{hardly be transferred} to different frameworks.
During the fine-tuning, adapter layers have to be \textit{empirically customized} and inserted into pretrained models, which introduces extra computation costs.
Therefore, we are motivated to ask our core question:

{\centering
\textit{Can we design \textbf{principled} methods to make OVS efficient, and seamlessly \textbf{transferable} to different frameworks?}
}

This work targets achieving performance as competitive as large vision-language foundation models but using \ul{smaller models with less training costs}.
We propose a Transferable Open-vocabulary segmentation technic (dubbed OpenTrans) , to establish a principled and seamlessly transferable efficiency across various OVS frameworks, eliminating the need for additional customization. This enables us to achieve the ``one design, transferable to all'' fashion.
\textbf{First}, to address the large model size of the CLIP image encoder, we explore small subnetwork by iterative magnitude pruning. To avoid overfitting the seen OVS classes and finding sparse masks specific to OVS frameworks, we prune the CLIP image encoder without semantic supervision.
This strategy enables our pruning approach to discover highly transferable subnetwork that seamlessly adapt to the latest OVS frameworks without requiring any modifications, avoiding any further pruning costs.
\textbf{Second}, to address the heavy OVS fine-tuning computation, we propose to only select partial layers to update.
To enable a ``plug-and-play'' strategy for layer-wise selective fine-tuning, we target analyzing the quality of pretrained layers without any data or any forward/backward process.
We investigate the spectrum of pretrained weights, freeze layers with heavy tails, and only update those with light tails.
This principled analysis not only further accelerates the training with fewer computations, but also introduces zero overhead in the pretrained model and can be directly adopted in different OVS frameworks.
We conduct comprehensive experiments on diverse OVS benchmarks, our method can employ {21.2M} smaller networks (Parameters), {59.11P} less training costs (Training FLOPs).

We summarize our contributions below:
\begin{itemize}
    \item Transferable sparse model: We reduce the backbone for {21.2M} smaller in parameters and reduce {94.6G} FLOPs. We avoid extra pruning costs on diverse latest OVS frameworks by directly transferring our subnetwork.
    \item Principled efficient fine-tuning: We design a plug-and-play layer-wise selection approach to enable efficient fine-tuning that reduces the training costs in OVS.
    \item Superior mIoU-efficiency trade-off: Extensive experiments demonstrate that we achieve comparable or even improved mIoU over diverse OVS benchmarks with significantly reduced computation costs for both training and inference.
\end{itemize}

\section{Related work}

\subsection{Vision-Language Pre-training Model}
Vision-language pre-training is a pre-training approach that integrates language and vision domains. In its early stages~\cite{li2020oscar, li2020unicoder}, research in this field was initially limited to smaller datasets, requiring subsequent fine-tuning on downstream tasks.
Recent studies emphasize the advantages of leveraging the extensive pool of web data accessible to researchers. Through the implementation of contrastive learning, CLIP~\cite{radford2021learning} establishes meaningful connections between images and their corresponding captions, achieving notable performance in cross-modal alignment. Alin~\cite{jia2021scaling} adopts a straightforward dual-encoder architecture and employs a contrastive loss to align the visual and linguistic representations of image and text pairs. The advent of these methods has played a substantial role in advancing Open-Vocabulary Segmentation.
{
Despite the promising performance, vision-language pre-trained models still impose significant computational overheads, primarily attributed to the large sizes of the backbone. This poses challenges in terms of wide applicability and affordability.
In contrast, we utilize CLIP as a source of open-domain capabilities, but propose efficiency to its image encoder with transferable sparse masks.
}

\subsection{Open-Vocabulary Segmentation}
The purpose of OVS is to segment diverse classes, including those that are inaccessible during the training process. The developmental trajectory of OVS can be delineated into two phases: pre- and post-introduction of vision-language pretraining models. Research conducted before the advent of vision-language pre-training models, as observed in works such as~\cite{bucher2019zero, xian2019semantic}, often yield suboptimal results. Therefore, integrating these models into OVS emerges as a highly advantageous strategy~\cite{liang2023open, jiao2024learning, ding2022open}.
Just after CLIP came out, Simbaseline~\cite{xu2022simple} adopts proposal masks for cropping the input image and leverages CLIP to extract region-level features. Zegformer~\cite{ding2022decoupling} utilizes CLIP as its encoder and incorporates MaskFormer~\cite{cheng2021per} for extracting mask proposals, both demonstrating improved performance in OVS. Recently, CAT-Seg~\cite{cho2023catseg} introduces a spatial and class aggregation network, integrating multi-modality guidance features to enhance open-vocabulary pixel classification effectiveness, achieving state-of-the-art performance on part datasets. FC-CLIP~\cite{yu2024convolutions} proposes a structure with two classifiers and freezes the CLIP backbone to uphold open-vocabulary recognition, also attaining state-of-the-art performance on certain datasets. Han et al.~\cite{han2023global} develop an efficient framework that avoids relying on the extra computational burden of the CLIP model.
{
In our research, our focus is on advancing principled efficiency for OVS frameworks. This is accomplished through the implementation of transferable sparse masks and efficient fine-tuning, employing spectrum analysis of pretrained weights.
}

\subsection{Parameter-Efficient Transfer Learning}
Parameter-Efficient Transfer Learning (PETL) is a research direction focused on minimizing the computational burden associated with adapting large pretrained models to new tasks. This is achieved by selectively updating specific parameters, eliminating the necessity to update the entire model. Adapters~\cite{rusu2016progressive,houlsby2019parameter} are compact bottleneck modules that are inserted within transformer layers.Empirical studies have demonstrated that using layer normalization layers alone for training adapters is adequate to achieve optimal performance during the fine-tuning process. Side-Adapters~\cite{xu2023side, sung2022lst} partitions the trainable parameters of the backbone model and forms a distinct side network. The designated task for this side network is to adapt the entire model to novel tasks, leading to its specific design tailored to different structures and tasks. Following a similar pattern, LoRA~\cite{hu2021lora} introduces trainable rank decomposition matrices into a pre-trained model that remains fixed. Although the majority of advancements in PETL have been observed in the NLP domain, researchers also extend the application of this technique to the fields of vision-language (VL)~\cite{zhou2022learning, zhang2021tip, sung2022vl, xu2023side}.
{
However, the adapter-based efficient fine-tuning methods still have to introduce extra layers or modules to the pretrained layers, which require extra design and customization efforts.
Instead, we aim to propose a principled and plug-and-play efficient fine-tuning method that can be seamlessly adopted in different OVS frameworks.
}


\begin{figure*}[h]
\centering
\begin{center}
   \hspace*{0.0cm}
   \includegraphics[width=0.95\linewidth]{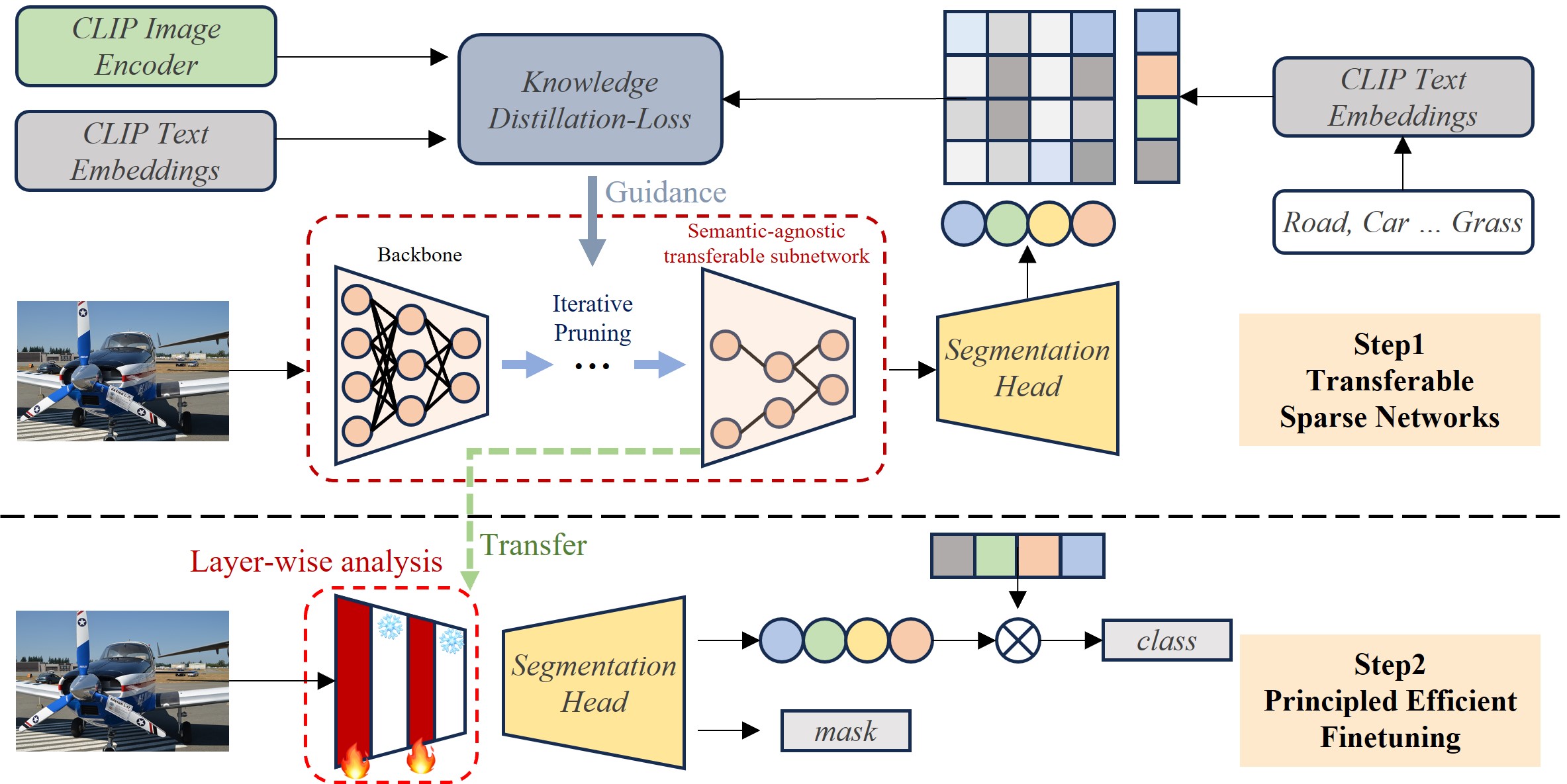}  
\end{center}
\vspace{-4mm}
\caption{Overview of OpenTrans(Ours). We introduce principled and transferable efficiency in two folds.
\textbf{Step1}: we prune the heavy CLIP image encoder without semantic awareness, this turns the backbone into a \textcolor[RGB]{192,0,0}{semantic-agnostic transferable subnetwork}. This allows us to seamlessly
transfer subnetwork to other OVS Frameworks,
such as Deeplabv3~\cite{chen2017rethinking}, Mask2former~\cite{cheng2022masked} and FC-CLIP~\cite{yu2024convolutions}.
\textbf{Step2}: during fine-tuning, we further explore and prioritize principled efficiency by introducing \textcolor[RGB]{192,0,0}{layer-wise heavy-tail spectrum analysis}. This method involves selectively updating layers with light-tail spectra in their pretrained weights, while keeping layers with heavy-tail spectra frozen. 
}
\vspace{-4mm}
\label{fig:structure}   
\end{figure*} 
\section{Method}\label{sec:method}

\subsection{Preliminaries}
\label{sec:method_background}

\paragraph{General OVS Pipeline.}

{OVS aims at training a segmentation model capable of segmenting arbitrary classes using text descriptions.
Most of the existing methods decoupled OVS into two steps: mask proposals generation and mask proposals classification. They first involve MaskFormer\cite{cheng2022masked} to generate class-agnostic mask proposals, followed by leveraging CLIP's transferability to classify these proposals. The mask proposals classification process can be defined as:
\begin{equation}
    C_i = v_i \mathop{\scriptstyle\bigotimes} T = softmax((v_i * t_1), (v_i * t_2) \cdots (v_i * t_{|C|})),\label{con:dot}
\end{equation}
where $C$ is the classification score. $v_i$ is the vision embeddings for $i$ mask proposals, which executed with either masked crops~\cite{ding2022decoupling, xu2022simple} or masked attention~\cite{xu2023side}. $T$ is the text embeddings from CLIP-pretrained text encoder. Building upon this inspiration, Our framework maintains the utilization of the Mask2Former architecture..}



During training, in order to preserve the generalizability of CLIP, we employ a distillation loss~\cite{han2023global} (``Knowledge Distillation Loss'' in ~\cref{fig:structure}, denoted as $\mathcal{L}_{\text{KD}}$) to distill the CLIP's knowledge to our backbone.
The distillation loss guarantees that the distance between text embeddings and visual embeddings remains consistent across all pairwise categories. The alignment of the CLIP text space with the image feature space enhances the performance of open vocabulary, while \ul{being agnostic to any semantic class}. Refer to the supplementary material for details.
In our work, we introduced this loss to enhance the open-domain characteristics of our pruning method.





\noindent\textbf{Out Target.}
In this work, we target introducing two levels of efficiency to current OVS works: both model and training efficiency.
More importantly, our core contribution lies in not only establishing these efficiency principles but also ensuring their transferability across various OVS frameworks. 
We will introduce two steps (see ~\cref{fig:structure}):
\begin{enumerate}
    \item \textbf{Model} efficiency (Sec.~\ref{sec:transfer}): We prune the heavy CLIP image encoder without semantic awareness to make the backbone transferable to different OVS frameworks.
    \item \textbf{Training} efficiency (Sec.~\ref{sec:efficient_finetune}): During fine-tuning, we only partially update selective layers of the backbone to further reduce training costs.
\end{enumerate}

\subsection{Model Efficiency}
\label{sec:transfer}
In this section, our main objective is to improve model efficiency through model pruning. Specifically, we aim to answer the question:
\textit{Can we find efficient subnetwork that can be seamlessly transferred to different OVS frameworks?}
We focus on finding a subnetwork within the model's backbone that allows for the seamless transfer of its sparse masks across different OVS works.




\noindent\textbf{Pruning method.} 
To identify the pruning mask for the backbone, we utilize the classical Iterative Magnitude Pruning (IMP) approach, as described in the LTH (Lottery Ticket Hypothesis) literature~\cite{chen2020lottery, chen2021lottery, frankle2018lottery, frankle2020linear}. The pruning of the model's backbone is carried out in a two-step process. First, we train an unpruned network as the pre-training model. Then, we remove a portion of weights with the globally smallest magnitudes~\cite{chen2021lottery, han2015deep, renda2020comparing}.

We acknowledge the crucial role of the knowledge distillation loss $\mathcal{L}_{\text{KD}}$ in aligning feature spaces between CLIP text and vision features, which significantly impacts the model's open-vocabulary performance. Hence, during the pruning process to find the pruning mask, we solely utilize the knowledge distillation loss and exclude any semantic supervision. This approach also allows us to remain agnostic to any specific OVS framework.
More importantly, it helps us discover a pruning mask on one OVS framework that can be transferable to other frameworks without any further customization, while still preserving the open-domain capabilities in CLIP without any risk of overfitting to specific semantic classes.


 We denote our backbone as $f(x;\theta)$, where $x$ is the input and $\theta \in \mathds{R}^{d}$ as its parameters. The subnetwork of the backbone can be described as $f(x;m\odot\theta)$ with a pruned binary mask $m\in\{0,1\}^{d}$, where $\odot$ is the element-wise product.
We depict our pruning outlines in ~\cref{alg:imp}, the sparsity level indicates the remaining weights of the model. \par

\begin{algorithm}[]
    \small
    \caption{Semantic-agnostic Model Pruning.}
    \begin{algorithmic}[1]
    \State Set the initial mask to $m = 1^{d}$, with pre-trained parameters $\theta$. 
    \Repeat
    \State Train $f(x; m \odot \theta)$ for $t$ iterations with only the distillation loss $\mathcal{L}_{\text{KD}}$.
    \State Prune $p\%$ of remaining weights in $\theta$ and update $m$ accordingly.
    \Until{the sparsity of $m$ reaches the desired sparsity level $s$.}
    \State Return $f(x; m \odot \theta)$.
    \end{algorithmic}
    \label{alg:imp}
\end{algorithm}

\begin{figure}[h]
\centering
\begin{center}
   \hspace*{-0.2cm}
   \includegraphics[width=1.0\linewidth]{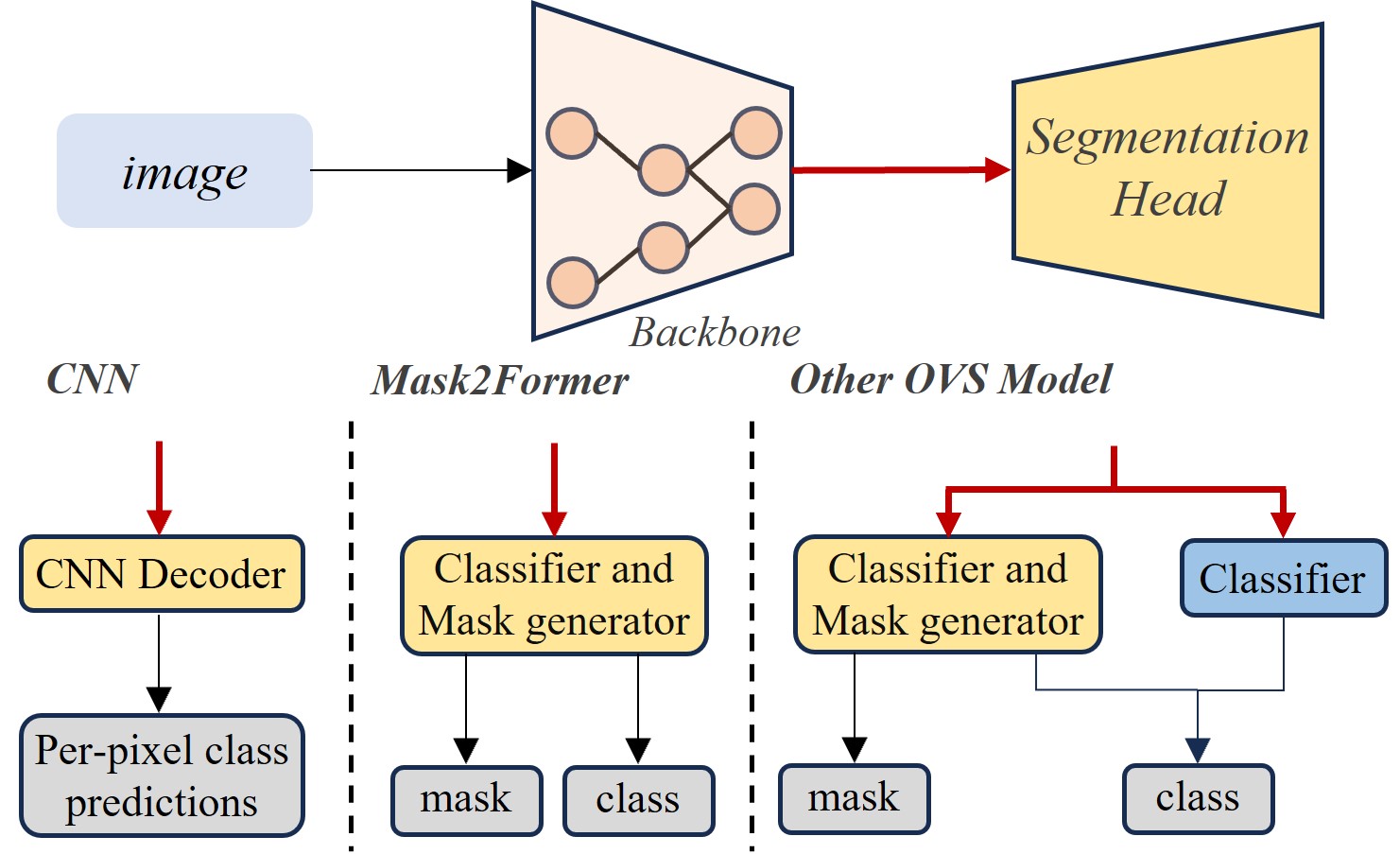}
\end{center}
\vspace{-4mm}
\caption{
We transfer our semantic-agnostic sparse masks to different OVS pipelines. \textbf{Left}: Traditional CNN-based frameworks: these frameworks commonly employ simple upsampling techniques. \textbf{Middle}: Mask2Former-based frameworks, the approach utilizes a decoupled mask and class segmentation head. \textbf{Right}: OVS with extra classifiers: some frameworks propose to leverage a frozen CLIP classifier to further utilize the pretrained visual knowledge (with the \textcolor{blue}{blue} box indicating frozen classifier layers of the model).
}
\vspace{-10pt}
\label{fig:sem_seg_head}   
\end{figure} 
\noindent\textbf{Subnetwork Transfer.} Upon obtaining the pruning mask through IMP, we proceed to transfer it to other OVS frameworks (\cref{fig:sem_seg_head}) by applying it to different segmentation backbones~\cite{yu2024convolutions, chen2017rethinking, cheng2022masked}. 
Importantly, no fine-tuning or additional adjustments are required when incorporating our subnetwork into these alternative frameworks, as it seamlessly integrates with the model.

To demonstrate that our transferable subnetwork can be widely adopted, we choose to study three representative and diverse OVS frameworks: 
{
\begin{itemize}
    \item Traditional CNN-based OVS: In the traditional convolutional segmentation model, CLIP text embeddings can be utilized as the classifier. Thus, in \cref{con:dot}, the $v_i$ represents the vision embeddings for the $i$-th pixel. We choose DeeplabV3 ~\cite{chen2017rethinking} as our transfer target, which incorporates the ASPP module to enhance its performance.
    \item Mask2Former-based OVS: This line of work separates the process of acquiring masks and predicting mask categories, resulting in higher quality masks compared to CNN methods. We choose Han et al.~\cite{han2023global} as our transfer target.
    \item OVS with extra classifiers: Beyond Mask2Former, many recent OVS frameworks additionally utilize an extra pre-trained classifier obtained from CLIP. We choose FC-CLIP~\cite{yu2024convolutions} as our target, which uses an out-of-vocabulary classifier during inference and greatly helps in OVS.  
\end{itemize}
}
In our experiments (Sec.~\ref{sec:transfer_exp}), we will show that by adopting our transferable subnetwork, we can facilitate the efficiency of all three types of OVS frameworks.
{After transferring the subnetwork, the sparsified pretrained backbone will be fine-tuned with CLIP text embeddings in the next section.}

\subsection{Training Efficiency} \label{sec:efficient_finetune}

{
In current OVS frameworks, heavy training costs are a common issue. The fine-tuning stage typically updates all layers without considering the quality of different pretrained weights, resulting in inefficiencies.
In this section, our objective is to address the question: \textit{Can we make the fine-tuning of OVS frameworks principled and efficient?}
To achieve this, we propose a principled and explainable metric that guides the fine-tuning of pretrained OVS models. Importantly, our approach does not introduce any extra parameters and does not incur any overhead during training.
}

\begin{figure}[h]
\centering
\begin{center}
   \includegraphics[width=1.0\linewidth]{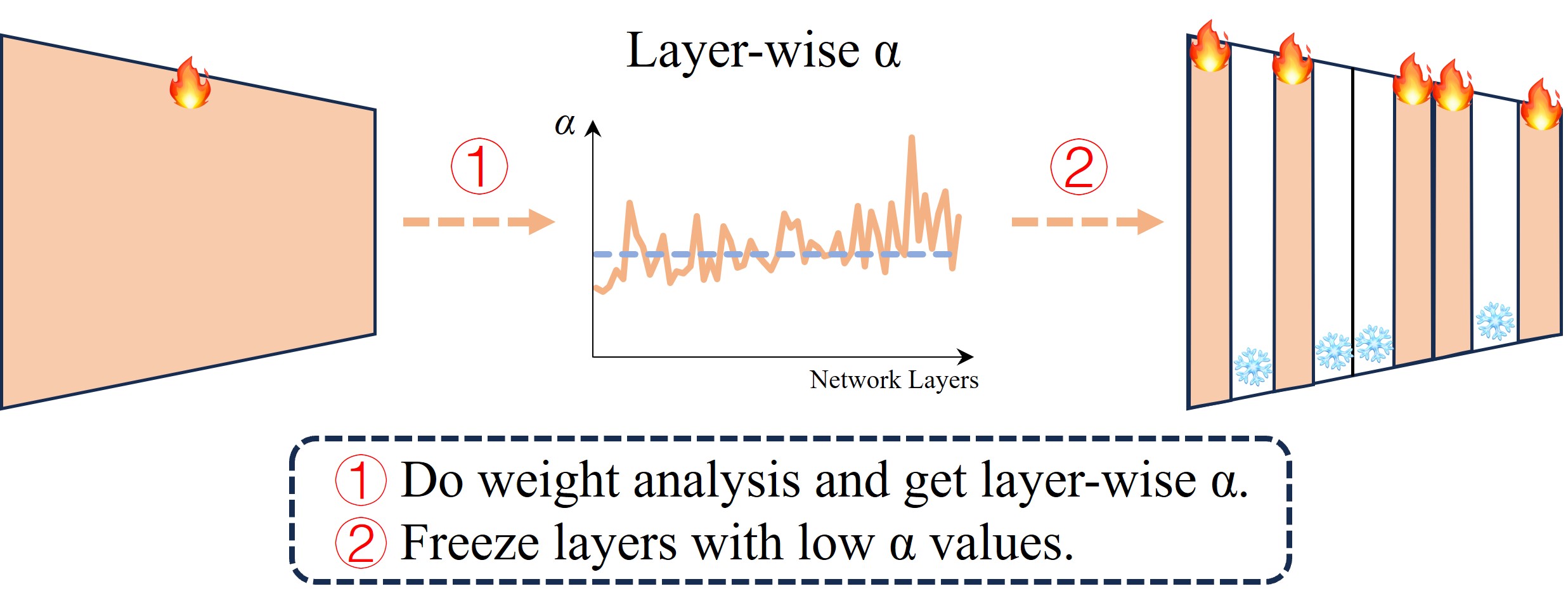}  
\end{center}
\vspace{-4mm}
\caption{We partially fine-tune layers of our pretrained backbone by analyzing heavy-tail spectrum of pretrained weights. We first fine-tune the whole model for $10^4$ iterations. Then, we compute $\alpha$ values for all layers of the model. During fine-tuning, we freeze layers with small $\alpha$ values (indicating good pretrained quality) and only fine-tune layers with large $\alpha$ (bad pretrained quality).}
\vspace{-4mm}
\label{fig:alpha}   
\end{figure} 

\subsubsection{Fine-tuning with Selective Layers} \label{sec:finetune_strategy}

We target a plug-and-play efficient fine-tuning, with minimal extra computation costs and without introducing any extra layers.
Our strategy is simple and principled: during fine-tuning process, we selectively calculate gradients and update layers that are still under-trained (weights of worse pretrained quality), while freeze and skip calculating gradients of layers that are already well-trained (weights of good pretrained quality), thus saving computation costs during fine-tuning.
We will defer how to determine the quality of pretrained layers in ~\cref{sec:heavy_tail}.
We depict our selective layer-wise fine-tuning in ~\cref{fig:alpha}.
\begin{table*}[t]
    \centering
    \small
    \renewcommand\arraystretch{1.0}
    \caption{\textbf{Model efficiency via subnetwork transfer}.
    By discovering semantic-agnostic sparse masks, we can directly transfer the subnetwork to different OVS frameworks, significantly reducing their model sizes and computation costs, while preserving their OVS performance after fine-tuning.
    $Random$ means trasnfering random subnetworks.
    }
    \vspace{-4pt}
    \setlength\tabcolsep{3pt}
    \resizebox{0.9\linewidth}{!}{
    \begin{tabular}{l|ccccccc|ll}
        \toprule[1pt]
        Model &COCO & Cityscapes& ADE20K-150 & ADE20K-847 & PAS-20 & PC-59 & PC-459 & Params. & FLOPs\\
        \hline
        Han et al.~\cite{han2023global} &46.0 &33.9 &16.6 &2.5 &71.2 &39.0 &7.1 &44.1M &268.2G\\
        $Random$ &37.4 &28.7 &13.2  &2.2 &60.1 &33.2 &5.8 &22.9M &173.3G\\
        Ours &42.5 &31.7 &15.8 &2.6 &64.6 &35.1 &6.4 &22.9M$_{\textcolor{red}{~48.1\%\downarrow}}$ &173.3G$_{\textcolor{red}{~35.4\%\downarrow}}$\\
        \hline
        DeeplabV3\cite{chen2017rethinking} &26.3 &20.3 &8.8 &- &44.1 &23.9 &4.1 &40.3M &241.3G \\
        $Random$ &17.9 &16.3 &6.4 &- &30.2 &16.5 &2.7 &19.1M &146.8G \\
        Ours &34.8 &24.3 &10.8 &- &55.2 &28.9 &5.2 &19.1M$_{\textcolor{red}{~52.6\%\downarrow}}$ &146.8G$_{\textcolor{red}{~39.2\%\downarrow}}$ \\
        \hline
        FC-CLIP~\cite{yu2024convolutions} &58.7 &53.2 &23.3 &7.1 &89.5 &50.5 &12.9 &39.0M &200.1G \\
        $Random$ &52.8 &50.0 &17.2 &3.2 &85.5 &44.8 &8.7 &17.8M &105.6G \\
        Ours &56.8 &52.1 &19.1 &4.2 &87.6 &47.4 &9.9 &17.8M$_{\textcolor{red}{~54.4\%\downarrow}}$ &105.6G$_{\textcolor{red}{~47.2\%\downarrow}}$ \\
        \bottomrule[1pt]
    \end{tabular}
    }
    \vspace{-10pt}
    \label{table:transfer}
    \end{table*}

\subsubsection{Layer Selections via Heavy-tail Analysis}
\label{sec:heavy_tail}

We assess the quality of pretrained weights and determine under-trained/well-trained layers by analyzing their \ul{heavy-tail behaviors} in the weights spectrums.

The fundamental concept of heavy-tail theory~\cite{martin2020heavy} is that in the empirical spectral density (ESD) of weight matrices, heavy-tail structures emerge as these matrices become more correlated and effectively capture diverse correlations in data during the optimization process~\citep{martin2018implicit_JMLRversion,martin2019traditional,martin2020heavy,martin2020predicting_NatComm,MM21a_simpsons_TR}. A key practical implication of this theory is the ability to forecast model performance by determining the power-law coefficient from the ESDs, which only requires the weights of the model.
Studies, such as Yang et al.~\cite{yang2022evaluating}, have shown that lower coefficients typically indicate greater test accuracy.

Imagine a neural network composed of $L$ layers, each associated with weight matrices $\Wb_1$, $\Wb_2$, $\cdots$, $\Wb_L$. For every weight matrix $\Wb_i$, which has a dimension of $N\times M$, assuming $N \geq M$, the correlation matrix is defined as $\bm{\Sigma}_i=\Wb_i^\top \Wb_i$. The eigenvalues of $\bm{\Sigma}_i$ are represented as $\{\lambda_j\}_{j=1}^M$, where $\lambda_j$ equals the square of the singular values of $\Wb_i$, denoted as $\{\sigma_j\}_{j=1}^M$. The largest eigenvalue of the correlation matrix $\bm{\Sigma}_i$ is referred to as $\lambda_{i,\max}$. The term ESD (empirical spectral density) for the weight matrix $\Wb_i$ describes the empirical distribution of the eigenvalues of $\bm{\Sigma}_i$, often visualized through a histogram. The density function that models the ESD, denoted as $p(\lambda)$, is considered within the range $(\lambda_{\min}, \lambda_{\max})$.
For a power law, $p$ satisfies
\begin{equation}\label{eqn:ALPHA}
    p(\lambda) \propto \lambda^{-\alpha}, \quad \lambda_{\min}<\lambda<\lambda_{\max}.
\end{equation}
$\alpha$ is essentially the slope of the tail of the ESD of the weights on a log-log scale.

A \textbf{smaller} $\alpha$ indicates a slow decay of large eigenvalues in the ESD, which tends to lead to a \textbf{better} quality of pretrained weights~\citep{martin2018implicit_JMLRversion}. Therefore, during fine-tuning, we only update layers with large $\alpha$ values (i.e. weights that are not well-trained), and will freeze layers with small $\alpha$ values (i.e. pretrained weights of good quality). {We update 50\% layers of top (larger) $\alpha$ values, and freeze the other 50\% layers of smaller $\alpha$ values, which helps prevent overfitting.}

\section{Experiments}

Sufficient experiments are conducted to verify the effectiveness of the model.
We first separately study the contribution from our transferable subnetworks (in Sec.~\ref{sec:transfer_exp}) and our principled layer-selective fine-tuning (in Sec.~\ref{sec:Layer-wise_finetune_experiments}).
We then show our final results (in Sec.~\ref{sec:final_results}) to demonstrate our superior balance between OVS accuracy and efficiency.

\subsection{Implementation Details}
\noindent \textbf{Datasets.} 
We conduct experiments in the following popular OVS datasets. \textbf{COCO-Panoptic} \cite{lin2014microsoft} is a large-scale semantic segmentation dataset with 133 classes. \textbf{ADE20K}~\cite{zhou2017scene} is a large-scale dataset for semantic segmentation, it contains more than 20k training images and 2k validation images. This dataset is divided into two splits. ADE20K-150 comprises 150 semantic classes, while ADE20K-857 encompasses 857 classes. \textbf{Cityscapes} ~\cite{cordts2016cityscapes}  is a dataset for urban scene understanding, its' validation set contains 500 high-resolution images. \textbf{PASCAL} datasets ~\cite{everingham2010pascal, mottaghi2014role} contains two splits of datasets. Pascal VOC 2012 contains 1,449 validation images from 20 classes. Pascal Context is an extensive of Pascal VOC 2010, and it has 5005 validation images. We utilize both the commonly used PC-59 version and the more challenging PC-459 version.

We evaluate our OpenTrans in a cross dataset setting following previous works~\cite{han2023global, yu2024convolutions, xu2023side, cho2023catseg}. Our subnetwork is trained on COCO and evaluated on the others. 

\noindent \textbf{Training Strategy.} Our implementation is inspired by the work of Han et al.~\cite{han2023global}. Throughout the model training process, we consistently utilize four NVIDIA RTX 4090 Ti GPUs. The input image is resized to $512\times512$. 
During the IMP stage outlined in Algorithm \ref{alg:imp}, we set the learning rate to 0.0001 without implementing any specific step schedule. The parameters $s$ and $p$ in the algorithm indicate a sparsity level and pruning ratio of 10\%, respectively.
When transferring the subnetwork to different architectures, we perform fine-tuning for several iterations. The experimental settings are adjusted accordingly, depending on the specific architectures being used. For a thorough understanding of the experimental details, please refer to the supplementary material.                               

\noindent \textbf{Parameters and FLOPs.} 
For precise and equitable computation, we employed a uniform calculation methodology to assess parameters and FLOPs across all tasks. Our evaluation encompasses all model parameters during testing. FLOPs were consistently computed using the initial graph from the COCO validation set with dimensions $800\times1216$.

\begin{table*}[t]
    \centering
    \small
    \renewcommand\arraystretch{1.0}
    \caption{\textbf{Efficient Fine-tuning}. By analyzing the heavy-tail spectrum in pretrained weights, our layer-selective strategy can reduce training costs in OVS frameworks in a principled manner. Here we choose Han et al.~\cite{han2023global} as our test-bed.  ``random'' indicates randomly freezing 50\% layers, ``$\alpha^*$'' means freezing 50\% layers with top (greater) $\alpha$ values, and ``$\alpha$'' means freezing 50\% layers with bottom (smaller) $\alpha$ values. 
    }
    \vspace{-4pt}
    \setlength\tabcolsep{3pt}
    \resizebox{0.9\linewidth}{!}{
    \begin{tabular}{l|ccccccc|l}
        \toprule[1pt]
        Traing method &COCO &Cityscapes &ADE20K-150 &ADE20K-847 &PAS-20 &PC-59 &PC-459 &Training FLOPs  \\
        \hline
        Han et al. &46.0 &33.9 &16.6 &2.5 &71.2 &39.0 &7.1 &181.4 P\\
        Han et al.(random) &44.5 &33.5 &16.4 &2.5 &70.1 &38.5 &7.2 &164.5 P $_{\textcolor{red}{~9.3\%\downarrow}}$\\
        Han et al.+$\alpha^*$  &45.3 &33.6 &16.7 &2.7 &73.2 &39.2 &7.3 &172.3 P $_{\textcolor{red}{~5.0\%\downarrow}}$\\
        Han et al.+$\alpha$ &\textbf{47.2} &\textbf{34.0} &\textbf{17.3} &\textbf{2.9} &\textbf{74.0} &\textbf{39.9} &\textbf{7.7} &\textbf{159.6 P}$_{\textcolor{red}{~12.0\%\downarrow}}$\\
        \bottomrule[1pt]
    \end{tabular}
    }
    \label{table:alpha_ablation}
    \end{table*}
\begin{table*}[t]
    \centering
    \small
    \renewcommand\arraystretch{1.1}
    \caption{\textbf{Balance: OpenSeg vs. Efficiency}. We compare OVS performance on popular image segmentation datasets.
    By adopting our methods, we can significantly boost the efficiency of popular OVS frameworks while preserving their OVS performance.
    }
    \vspace{-4pt}
    \setlength\tabcolsep{4pt}
    \renewcommand\arraystretch{1}
    \resizebox{0.9\linewidth}{!}{
    \begin{tabular}{l|ll|cccc|cc}
        \toprule[1pt]
        Model &Backbone &Training Set &COCO &PAS-20 &PC-59 &ADE20K-150  & Params. & FLOPs\\
        \hline
        ZS3Net ~\cite{bucher2019zero} &R-101 &PASCAL-15 &- &38.3 &19.4 &- &- &- \\
        SPNet ~\cite{xian2019semantic} &R-101 &PASCAL-15 &- &18.3 &- &24.3 &- &- \\
        LSeg ~\cite{li2022languagedriven} &R-101 &PASCAL-15 &- &47.4 &- &- &- &- \\
        ZegFormer ~\cite{ding2022decoupling} &R-50 &COCO Stuff &- &80.7 &42.8 &16.4 &63.9M &1128.7G\\
        Cat-Seg ~\cite{cho2023catseg} &R-101 &COCO Stuff &- &93.7 &57.5 &27.2  &60.8M &3847.8G\\
        OVSeg ~\cite{liang2023open} &R-101 &COCO Stuff &- &89.2 &53.3 &24.0 &61.1M &1116.2G\\  
        LSeg+ ~\cite{ghiasi2022scaling} &R-101 &COCO Panoptic &- &59.0 &36.0 &13.0 &- &- \\
        OpenSeg ~\cite{ghiasi2022scaling} &R-101 &COCO Panoptic &36.9 &- &36.9 &15.3  &- &-\\
        Simbaseline ~\cite{xu2022simple} &R-50 &COCO Panoptic &39.5 &- &40.1 &14.4 &64.6M &1060.6G\\
        DeeplabV3 ~\cite{chen2017rethinking} &R-50 &COCO Panoptic &26.3 &44.1 &23.9 &8.8 &40.3M &241.3G\\
        Han et al. ~\cite{han2023global} &R-50 &COCO Panoptic &46.0 &71.2 &39.0 &16.6 &44.1M &268.2G\\
        FC-CLIP ~\cite{yu2024convolutions} &R-50 &COCO Panoptic &58.7 &89.5 &50.5 &23.3 &39.0M &200.1G\\
        \hline
        Ours (Han et al.) &R-50 &COCO Panoptic &42.3 &64.8 &35.7 &15.3   &22.9M &173.3G\\
        Ours (DeeplabV3) &R-50 &COCO Panoptic &34.2 &54.6 &28.8 &10.7   &19.1M &146.8G\\
        Ours (FC-CLIP) &R-50 &COCO Panoptic &56.8 &87.3 &47.4 &18.8   &17.8M &105.6G\\
        
        
        \bottomrule[1pt]
    \end{tabular}
    }
    \vspace{-10pt}
    \label{table:performance}
    \end{table*}


\subsection{Model Efficiency via Subnetwork Transfer}
\label{sec:transfer_exp}
{
In this section, we employ three exemplary methodologies \cite{han2023global, chen2017rethinking, yu2024convolutions} to evaluate the efficiency of the transferable subnetwork transfer strategy. We compare these three representative methods with original approaches and random subnetwork transfer strategies.}

\noindent \textbf{Comparisons with original OVS approaches.}
We prune once but contribute to the efficiency of peer works via  subnetwork transfer. Application of our subnetwork yields notable reductions in Params and FLOPs for models such as Han et al., DeeplabV3, and FC-CLIP, as delineated in ~\cref{table:transfer}. This means that all our three models share the same subnetwork, which validates the transferability of our method. Specifically, for Han et al., we achieve a reduction  in Params from 44.1M to 22.9M, and FLOPs from 268.2G to 173.3G. Despite being nearly half the size, the mIoU only saw a marginal 2.2\% reduction in Cityscapes and 0.8\% in ADE20K-150.
In the case of DeeplabV3, we successfully halved the model size while surpassing its original performance. The improvements amounted to a 4\% increase in mIoU for Cityscapes and a 2\% boost for ADE20K-150.
Compared with the original FC-CLIP model, we are able to compress it to 17.95M Params and 105.6G FLOPs, surpassing even the compactness achieved by Han et al. Although FC-CLIP emphasizes the significance of freezing the CLIP backbone, it is noteworthy that the subnetwork we utilized exerted a minimal impact on the model's performance in the task of OVS.

\noindent \textbf{Comparisons with random subnetwork transfer.}
{The results for the $Random$, characterized by equivalent sparsity to our transferable subnetwork, are omitted in \cref{table:transfer}.
It is evident that ours significantly outperforms $Random$ across all datasets, This highlights the effectiveness of our proposed method in preserving the transferability of performance from previous OVS approaches. } 

In a word, the proposed transferable subnetwork significantly decreases computational redundancy with minimal impacts on performance.

\subsection{Efficient Fine-tuning with Layer Selections}
\label{sec:Layer-wise_finetune_experiments}
Based on the computation method outlined in ~\cref{sec:efficient_finetune}, we can compute the $\alpha$ value for each layer of the backbone. The model training is initiated for $10^4$ iterations before performing the $\alpha$ derivation and freezing specific layers.

In ~\cref{table:alpha_ablation}, we adopt diverse approaches to freeze half of the layers within the backbone. Through the identification of layers with small $\alpha$ values as high-quality layers, excluding them from training results in enhanced performance not only on the training dataset but also significant improvements compared to the original model across all evaluated OVS datasets. 
{Notably, we achieve 0.7\%, 2.8\%, and 0.9\% improvement on ADE20K-150, PAS-20 and PC-59 in terms of mIoU.}
In comparison to randomly freezing half of the layers without training involvement, our proposed method consistently surpasses the performance of the original model on all datasets. The freezing method allows for the conservation of training FLOPs, leading to more efficient resource utilization, as detailed in ~\cref{table:alpha_prune}. It becomes apparent that through fine-tuning the identified poorly performing layers, the network can attain improved training results while utilizing fewer parameters.

\subsection{Stronger Balance: OpenSeg vs. Efficiency}\label{sec:final_results}

To assess the open-vocabulary performance of our proposed method, we compare its effectiveness against state-of-the-art techniques on various well-known image segmentation datasets~\cite{lin2014microsoft, zhou2017scene, cordts2016cityscapes, everingham2010pascal, mottaghi2014role}.  
The results presented in ~\cref{table:performance} illustrate the outcomes of employing our two previously proposed methods for enhancing model and training efficiency. These findings showcase that our model not only attains impressive speed but also demonstrates robust OVS capabilities.
To elaborate, our method achieves a notable 56.8\% mIoU on the training set COCO, encompassing 133 categories. It further attains an 18.8\% mIoU on the ADE20K dataset with 150 categories and secures 47.4\% and 9.9\% mIoU on Pascal Context with 49 categories and 459 categories, respectively.

In comparison to the state-of-the-art method Cat-Seg across multiple datasets, our approach reveals a performance gap of only 6.4\% mIoU on the PAS-20 dataset and 8.4\% mIoU on the ADE20k-150 dataset. Notably, we achieve a remarkable reduction of over 240\% in model parameters, coupled with a nearly 40-fold improvement in computational speed. Han et al. pioneer the elimination of the two-stage process, demonstrating commendable performance and efficiency. In contrast, our approach showcases advantages of 2.2\% mIoU on ADE20K-150 and 8.4\% mIoU on PC-59. Moreover, we achieve more than a twofold improvement in both parameter efficiency and operational speed.

\begin{table}[t]
    \centering
    \small
    \caption{Reduction of training FLOPs with our methods (ResNet-50). ``$pruning$'' means our transferable subnetworks in Sec~\ref{sec:transfer}. ``$\alpha$'' means selective layer-wise fine-tuning in Sec~\ref{sec:efficient_finetune}.
    }
    \vspace{-4pt}
    \setlength\tabcolsep{3.5pt}
    \renewcommand\arraystretch{1.2} 
    \resizebox{0.7\linewidth}{!}{
    \begin{tabular}{l|l}
        \toprule[1pt]
        Model  &Training FLOPs\\
        \hline
        Han et al. &181.4 P\\
        Han et al.+$\alpha$ &159.6 P$_{\textcolor{red}{~12.0\%\downarrow}}$\\
        Han et al.+$\alpha$+$pruning$ &122.2 P$_{\textcolor{red}{~32.6\%\downarrow}}$\\
        \bottomrule[1pt]
    \end{tabular} \label{table:alpha_prune}
    }
    \vspace{-10pt}
\end{table}

\subsection{Training cost}
The layer-wise fine-tuning method we proposed, has proven to be effective across OVS works. 
Moreover, the combination of this layer-wise fine-tuning method with our subnetwork further significantly amplifies training efficiency. In ~\cref{table:alpha_prune}, our focus is exclusively on the training FLOPs of the entire model, given that our subnetwork and layer-wise fine-tuning methods are specifically applied to the backbone component. The incorporation of $\alpha$ results in a substantial reduction of 12.0\% in the training parameters for the model. When coupled with subnetwork, the training parameters are further diminished to only 32.6\% of their original value. A comprehensive breakdown of the training FLOPs calculation will be provided in the appendix section.

\vspace{2pt}
\subsection{Qualitative Results}
\label{sec:qualitative_results}
\begin{figure}[t]
\centering
\begin{center}
\vspace{-0mm}
   \hspace*{-0.2cm}
   \includegraphics[width=0.99\linewidth]{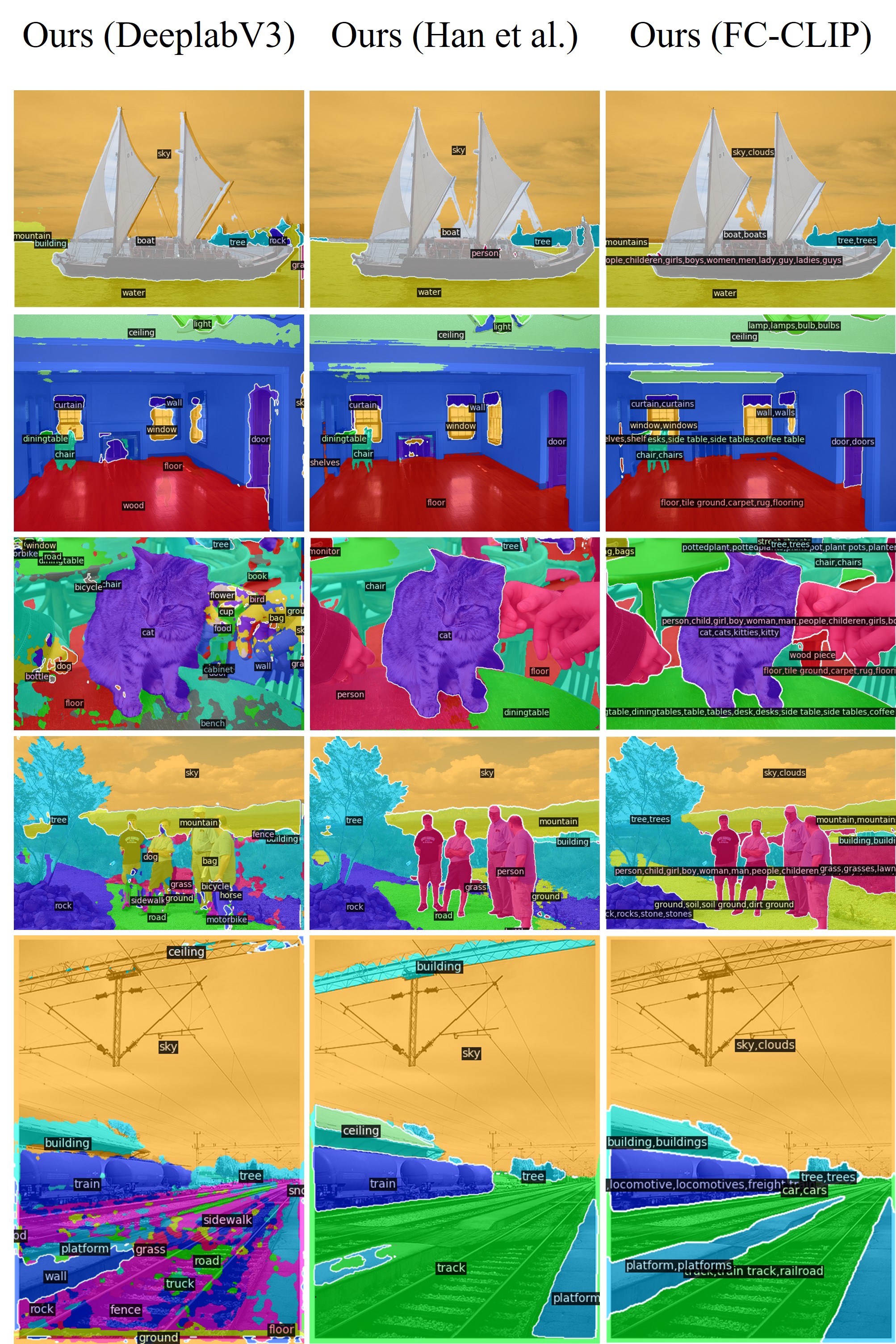}  
\end{center}
\vspace{-2mm}
\caption{Visualizations of examples on the PC-59 validation set by our models (trained on COCO panoptic training set, zero-shot evaluated on the PC-59 validation set).}
\vspace{-2mm}
\label{fig:vis}   
\end{figure}

\cref{fig:vis} displays several visualization results obtained from our method. 
There exists a notable disparity in the quality of masks obtained by various baselines. In intricate scenes, the conventional convolution segmentation head exhibits subpar performance. Moreover, the unique structural design of FC-CLIP facilitates a more precise assignment of unseen category labels to the generated masks, enhancing its capacity to predict both foreground and background categories.

\section{Conclusion}
In this paper, we introduce a transferable subnetwork approach that markedly reduces the size and computational demands of popular OVS models.
Additionally, we propose a principled layer-selective fine-tuning method, enabling efficient fine-tuning of OVS models by only updating under-trained layers. Through extensive experiments on various OVS datasets, we illustrate that our model strikes a commendable balance between performance and efficiency.  We hope our method can serve as a solid baseline and help advance the future research of OVS.
Some current limitations and potential future works: 1) adaptation of our method on larger convolutional backbones (e.g. ConvNeXt-Large) and ViT-based backbones; 2) more fine-grained selection of weight elements or channels for fine-tuning, going beyond the layer level; 3) extension to other open-vocabulary tasks, such as open-set object detection.

{
    \small
    \bibliographystyle{ieeenat_fullname}
    \bibliography{main}
}

\clearpage
\setcounter{page}{1}
\maketitlesupplementary

\section{Experimental Details}
\subsection{Fine-tuning Settings After Transfer}
\label{sec:sup_transfer_detail}
This section provides a detailed introduction to the experimental settings for all three transfer experiments. The experimental settings used for Han et al. and Deeplabv3 are identical. We trained the models on the COCO panoptic dataset for 50,000 iterations using four 4090 Ti GPUs. The training was performed with a batch size of 28, utilizing the SGD optimizer with an initial learning rate of 0.00015. The learning rate was reduced by a factor of 0.1 at 40,000 iterations and 45,000 iterations. For FC-CLIP, we conducted training for 368,750 iterations using eight 4090 Ti GPUs. The training was performed with a batch size of 16, utilizing the AdamW optimizer with an initial learning rate of 0.0001. The learning rate was reduced by a factor of 0.1 at 327,778 iterations and 355,092 iterations.

\subsection{Training FLOPs Calculation}
To calculate the training FLOPs of our model backbone, we consider both the forward and backward processes.
When calculating the backward FLOPs, we need to compute gradients for both model parameters and the hidden features. Hence, the FLOPs involved in the backward process will be doubled compared with the FLOPs of the forward process\footnote{\href{https://epochai.org/blog/backward-forward-FLOP-ratio}{https://epochai.org/blog/backward-forward-FLOP-ratio}}.
Suppose the inference FLOPs of the model is $C$.
We provide a comprehensive calculation approach below where our subnetwork has a sparsity of 10\% and 50\% of the layers are frozen during fine-tuning:
\begin{enumerate}
    \item \textbf{Standard Fine-tuning:} forward FLOPs 1$\times C$, backward FLOPs 2$\times C$.
    \item \textbf{Model with layer-wise fine-tuning:} \ul{Frozen layers}: forward FLOPs 1$\times C$, backward FLOPs 1$\times C$. \ul{Active layers}: forward FLOPs 1$\times C$, backward FLOPs 2$\times C$.
    \item \textbf{Model with subnetwork: } forward FLOPs 0.1$\times C$, backward FLOPs 1.1$\times C$ (1 for hidden features, 0.1 for sparse weights).
    \item \textbf{Model with layer-wise fine-tuning and subnetwork: } \ul{Frozen layers}: forward FLOPs 0.1$\times C$, backward FLOPs 1$\times C$. \ul{Active layers}: forward FLOPs 0.1$\times C$, backward FLOPs 1.1$\times C$ (1 for hidden features, 0.1 for sparse weights).
\end{enumerate}
To calculate the FLOPs of specific layers, we utilize the ``$get\_model\_complexity\_info$'' function provided by the $ptFLOPs$ library\footnote{\href{https://pypi.org/project/ptflops/}{https://pypi.org/project/ptflops/}}.

\begin{figure*}[h]
\centering
\begin{center}
   \hspace*{-0.1cm}
   \includegraphics[width=1.0\linewidth]{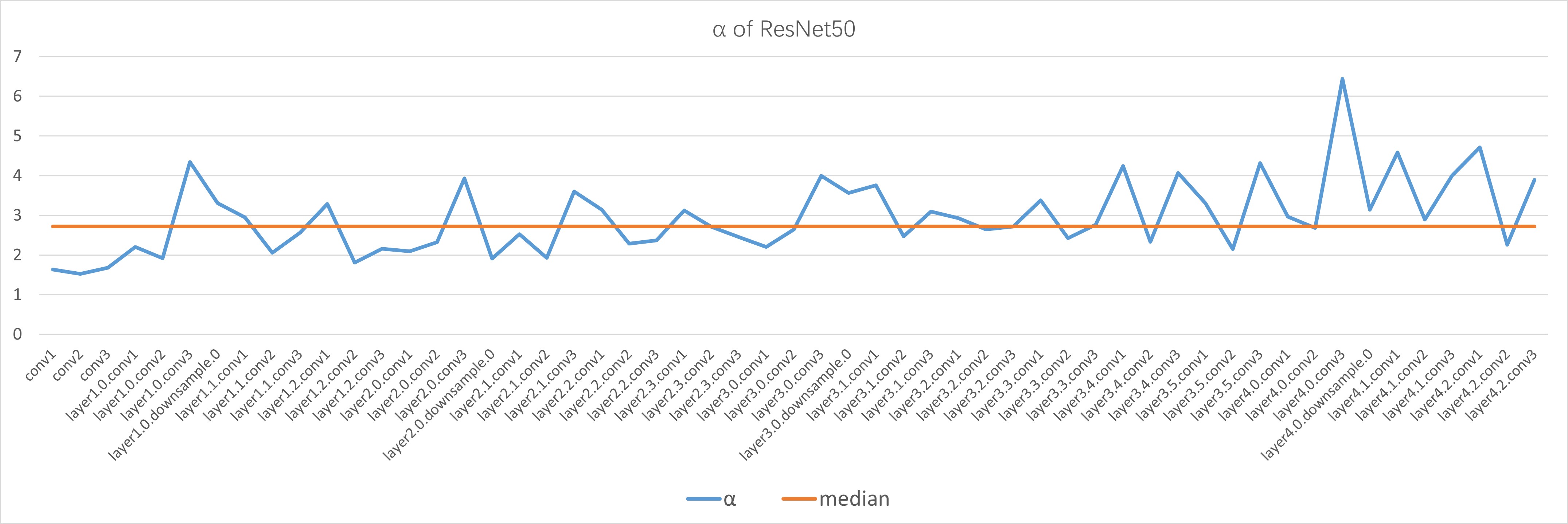}  
\end{center}
\vspace{-4mm}
\caption{Layer-wise $\alpha$ values in model backbone before fine-tuning used in ~\cref{sec:Layer-wise_finetune_experiments}. Specifically, the blue curve represents the calculated $\alpha$ value for each layer of Resnet, while the red line represents their median. During the fine-tuning process, we freeze the layers that have values smaller than the median.}
\vspace{-0mm}
\label{fig:sup_perlayer_alpha}   
\end{figure*} 

\subsection{Visualizations of Heavy-tail Behaviors in Efficient Fine-tuning}
\begin{figure}[h]
\centering
\begin{center}
   \hspace*{-0.1cm}
   \includegraphics[width=0.95\linewidth]{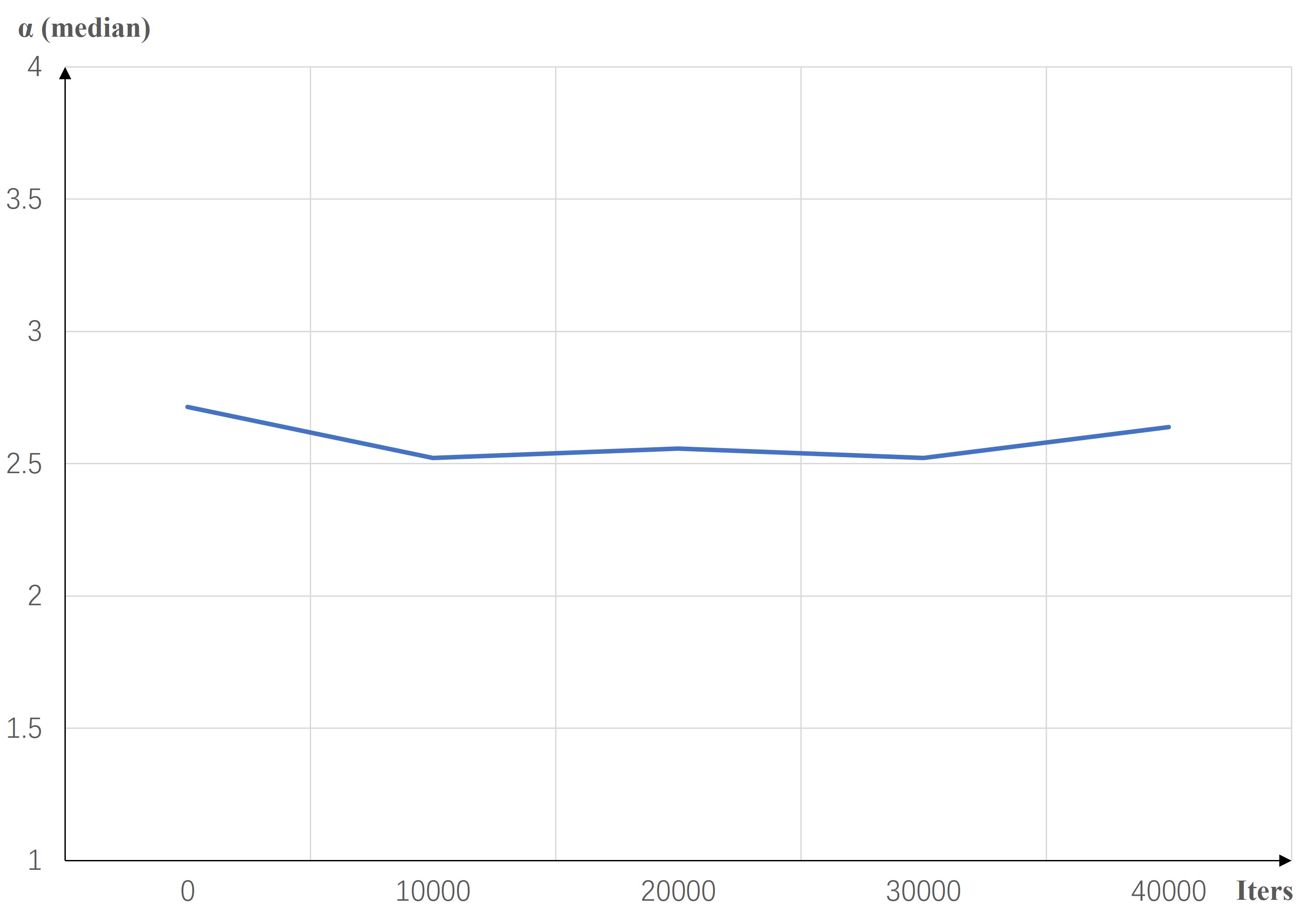}  
\end{center}
\vspace{-4mm}
\caption{The median value of layer-wise $\alpha$ during the fine-tuning process. We can see $\alpha$ remains relatively stable without significant changes.}
\label{fig:sup_pl_during_tuning}   
\end{figure}

Our ~\cref{fig:alpha} provides a comprehensive overview of the layer-wise fine-tuning implementation process.
{~\cref{fig:sup_perlayer_alpha} showcases the $\alpha$ value for each layer in our backbone before fine-tuning (after pruning, before adopting sparse masks). Layers of $\alpha$ below the red horizontal line (median value of $\alpha$ across layers) will be frozen during fine-tuning.}
Additionally, ~\cref{fig:sup_pl_during_tuning} illustrates the changes in the median of $\alpha$ values throughout the entire fine-tuning process under our training method. It is evident that the $\alpha$ value of most layers remains relatively stable, leading us to not dynamically calculate $\alpha$ or adjust our fine-tuning layers in the main experiment.

\section{Futher Experiments}
\subsection{Ablation Study on Different Knowledge Distillation Losses}
\label{sec:Knowlegde_Distillation_Loss_Ablation}
In the method presented in \cref{sec:method_background}, a loss function is introduced to align the text with the vision feature space. This loss, referred to as text-guided knowledge distillation (TGKD), facilitates the distillation of knowledge from textual information into visual embeddings. The process of text-guided knowledge distillation can be formulated as follows:
\begin{equation}\small
    \mathcal{L}_{TGKD} = \frac{1}{N}\sum_{i=1}^N\sum_{j=1}^N\Big\|(\|\mathcal{V}_i - \mathcal{R}(I, M_j)\| - \|\mathcal{T}(Y_i) - \mathcal{T}(Y_j)\|)\Big\|,
    \label{eq:tgkd}
\end{equation}
where $\mathcal{T}$ denotes the CLIP text encoder, $\mathcal{R}$ denotes the CLIP image encoder, $\mathcal{V}_i$ denotes the visual embeddings and $Y_i$ is the category name of $i$-th ground truth region. $I$ represents the input image, while $M$ represents the mask corresponding to the respective category.
Correspondingly, a vision-guided knowledge distillation approach has also been proposed, which can be formulated as follows:
\begin{equation}\small
    \mathcal{L}_{VGKD} = \frac{1}{N}\sum_{i=1}^N\sum_{j=1}^N\Big\|(\|\mathcal{V}_i - \mathcal{R}(I, M_j)\| - \|\mathcal{V}_i - \mathcal{V}_j\|)\Big\|,
\end{equation}
~\cref{table:sup_loss_ablation} provides a comprehensive comparison of the results obtained from pruning using two distinct distillation losses. The analysis demonstrates that the utilization of text-guided distillation loss significantly enhances the OVS performance of the model.
\begin{table}[t]
    \centering
    \small
    \caption{Comparison of different distillation loss: $\mathcal{L}_{TGKD}$ vs. $\mathcal{L}_{VGKD}$.}
    \label{tab:kd_strategy}
    \setlength\tabcolsep{3pt}
    \begin{tabular}{l|ccc}
        \toprule[1pt]
        Distillation Loss  &COCO &PC-59 &ADE20k-150\\
        \hline
        Text-guided ($\mathcal{L}_{TGKD}$) &42.5 &35.1 &15.8\\
        Vision-guided ($\mathcal{L}_{VGKD}$) &40.0 &32.6 &14.2\\
        \bottomrule[1pt]
    \end{tabular} \label{table:sup_loss_ablation}
\end{table}

\begin{table}[t]
    \centering
    \small
    \renewcommand\arraystretch{1.0}
    \caption{Ablation study of different configurations for pruning with both distillation and segmentation loss. Pruning ratio ($p$) and the number of training iterations ($t$) used in~\cref{alg:imp} were studied.
    }
    \setlength\tabcolsep{3pt}
    \begin{tabular}{cc|ccccc}
        \toprule[1pt]
        $p$ &$t$ &COCO &Cityscapes &ADE20K-150 &PAS-20 &PC-59 \\
        \hline
        0.1 &5000 &40.6 &29.1 &15.0 &60.1 &33.0 \\
        0.1 &2500 &36.9 &28.9 &13.8 &54.8 &32.0 \\
        0.2 &5000 &32.1 &27.8 &11.6 &50.4 &27.8 \\
        \bottomrule[1pt]
    \end{tabular}
    \vspace{2pt}
    \label{table:straight_prune_ablation}
    \end{table}
\begin{table}[t]
    \centering
    \small
    \renewcommand\arraystretch{1.0}
    \caption{
    Ablation study of different configurations for pruning with only distillation loss followed by segmentation fine-tuning (our strategy in Sec.~\ref{sec:method}). Pruning ratio ($p$) and the number of training iterations ($t$) used in~\cref{alg:imp} were studied.
    }
    \setlength\tabcolsep{3pt}
    \begin{tabular}{cc|ccccc}
        \toprule[1pt]
        $p$ &$t$ &COCO &Cityscapes &ADE20K-150 &PAS-20 &PC-59 \\
        \hline
        0.1 &5000 &41.8 &31.8 &15.4 &63.2 &35.1 \\
        0.1 &1000 &41.9 &32.7 &15.0 &63.9 &35.1 \\
        0.1 &100 &42.4 &31.0 &14.3 &63.2 &34.8 \\
        0.3 &1000 &40.4 &28.9 &13.7 &61.4 &33.6 \\
        0.5 &1000 &39.8 &30.3 &13.6 &60.7 &33.3 \\
        \bottomrule[1pt]
    \end{tabular}
    \vspace{-15pt}
    \label{table:prune_ablation}
    \end{table}
\begin{figure*}[h]
\centering
\begin{center}
   \hspace*{-0.2cm}
   \includegraphics[width=0.9\linewidth]{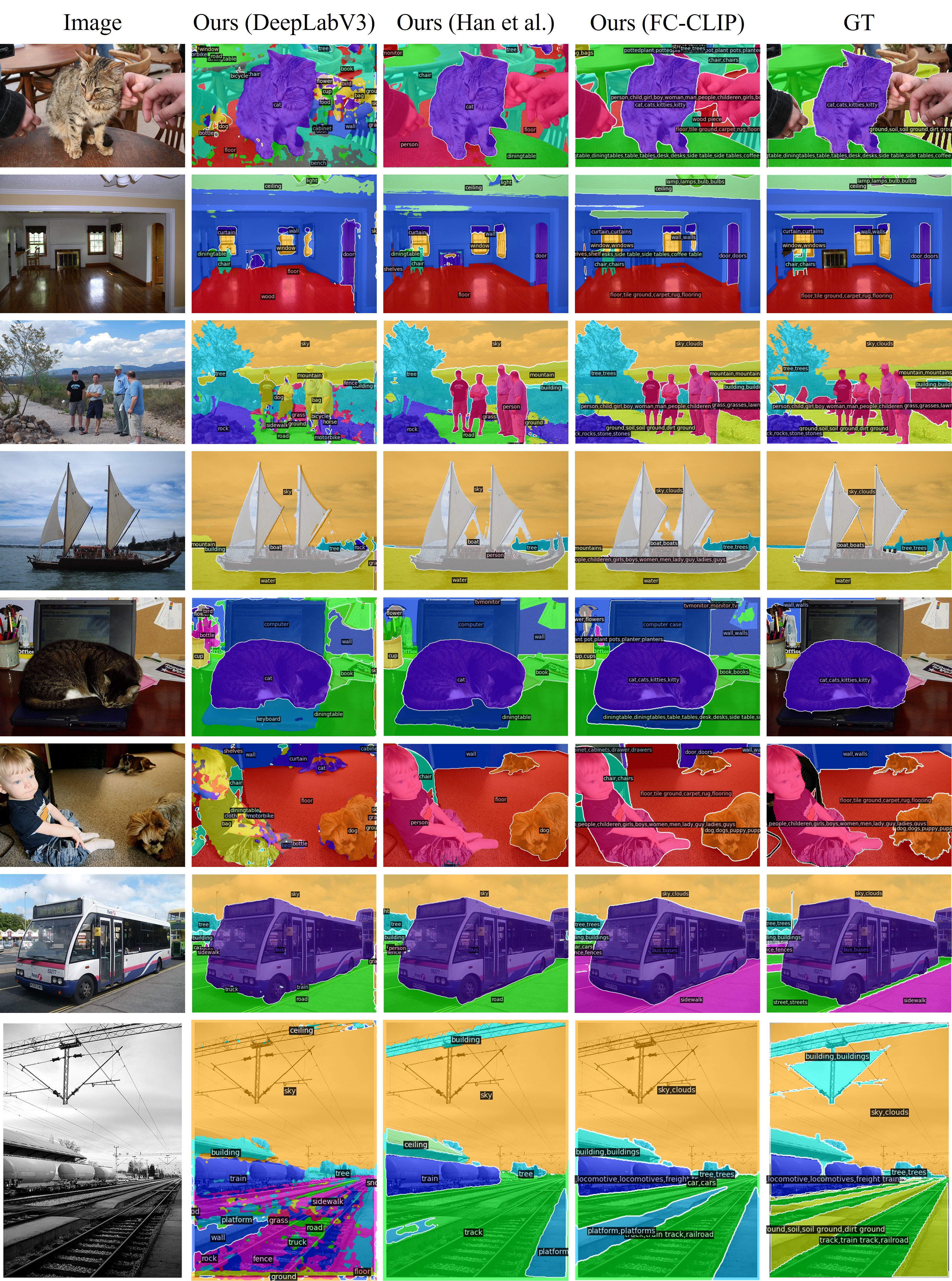}  
\end{center}
\vspace{-4mm}
\caption{More visualizations of examples on the PC-59. ``GT'': ground truth.}
\vspace{-2mm}
\label{fig:sup_vis}   
\end{figure*}

\subsection{Ablation Study on Pruning Strategies}
We further study two pruning strategies.
The first strategy is to naively apply IMP to prune the model using all training losses ({distillation loss plus mask loss and classification loss}).
The second strategy, which is proposed in our ~\cref{sec:transfer}, first obtains subnetworks based on the semantic-agnostic distillation loss, and subsequently fine-tunes the model for a specific number of iterations.

~\cref{table:sup_purning_method} provides experimental details. The results demonstrate that the adopted pruning with fine-tuning method not only improves OVS performance but also requires fewer training iterations. Additionally, the subnetworks obtained without semantic supervision exhibit the ability to be further fine-tuned in the original model and can also be effectively transferred to other tasks, yielding superior experimental results (as shown in ~\cref{table:transfer}).

\paragraph{Ablation Study on Pruning Configurations.}
We also study the best configuration of $t$ and $p$ of two pruning strategies discussed above.
Ablation studies on the first pruning strategy (distillation
+ segmentation loss) are shown in~\cref{table:straight_prune_ablation}.
It is important to highlight that reducing the value of $t$ or increasing the value of $p$ has a substantial impact on the performance, leading to a noticeable drop.

Additionally, for our pruning strategy (pruning with distillation only, followed by segmentation fine-tuning), different configurations are studied in ~\cref{table:prune_ablation}.
Initially, a pruning rate ($p$) of 0.1 was used, while different values of $t$ were employed.
The performance on the COCO dataset was found to be very similar across different $t$ values. However, when $t$ was reduced to a very small value, a significant drop in performance on the OVS datasets was observed. This observation further supports the claim that the subnetwork discovery method proposed in ~\cref{sec:transfer} is advantageous for the OVS task. 
If $t$ is kept constant and the pruning rate is increased, it will result in IMP finding subnetwork with lower quality. It is worth noting that the final sparsity of the subnetworks discovered by different pruning rates will vary, making direct comparisons between them relatively unfair. However, this disparity is inevitable given the different pruning rates employed.

\begin{table}[t]
    \centering
    \small
    \caption{We compare the segmentation performance (mIoU) of two strategies for finding sparse models: 1) pruning with both distillation and segmentation loss; 2) our pruning (with only distillation loss) followed by segmentation fine-tuning (Sec.~\ref{sec:method}). The ``Training Iters'' parameter represents the total number of training iterations required by each of the two methods.}
    \vspace{-4pt}
    \setlength\tabcolsep{2pt}
    \resizebox{\linewidth}{!}{
    \begin{tabular}{l||ccc||c}
        \toprule[1pt]
        Pruning Method &COCO &PC-59 &ADE20k-150 &Training Iters.\\
        \midrule
        \makecell{Pruning (distillation \\ + segmentation loss)} &40.6 &33.0 &15.0 &105000\\
        \hline
        \makecell{Pruning (distillation only) \\ + Segmentation Fine-tuning} &42.5 &35.1 &15.8 &95000\\
        \bottomrule[1pt]
    \end{tabular} \label{table:sup_purning_method}
    }
\end{table}
\begin{table}[t]
    \centering
    \small
    \caption{CKA similarity between CLIP image encoder and DeeplabV3 backbones.
    }
    \vspace{-4pt}
    \setlength\tabcolsep{3pt}
    \resizebox{0.8\linewidth}{!}{
    \begin{tabular}{l|cc}
        \toprule[1pt]
         &DeeplabV3 & Sparse DeeplabV3 (Ours)  \\
        \hline
        CKA (vs. CLIP) &0.361 &0.512 \\
        \bottomrule[1pt]
    \end{tabular}
    }
    \label{table:cka_ablation}
    \end{table}
\begin{table}[t]
    \centering
    \small
    \renewcommand\arraystretch{1.0}
    \caption{Ablation study of freezing ratio
    }
    \vspace{-4pt}
    \setlength\tabcolsep{3pt}
    \resizebox{1.0\linewidth}{!}{
    \begin{tabular}{l|ccccccc}
        \toprule[1pt]
        Ratio &COCO &Cityscapes &ADE-150 &ADE-847 &PAS-20 &PC-59 &PC-459  \\
        \hline
        0.25 &47.5 &34.5 &17.3 &2.9 &72.5 &39.6 &7.6 \\
        0.75  &46.5 &34.8 &17.4 &2.8 &72.6 &38.9 &7.5 \\
        0.5 &{47.2} &{34.0} &{17.3} &{2.9} &{74.0} &{39.9} &{7.7} \\
        \bottomrule[1pt]
    \end{tabular}
    }
    \vspace{-4pt}
    \label{table:alpha_ratio_ablation}
    \end{table}
\vspace{-4mm}
\begin{figure}[t]
\centering
\begin{center}
   \hspace*{-0.2cm}
   \includegraphics[width=1.0\linewidth]{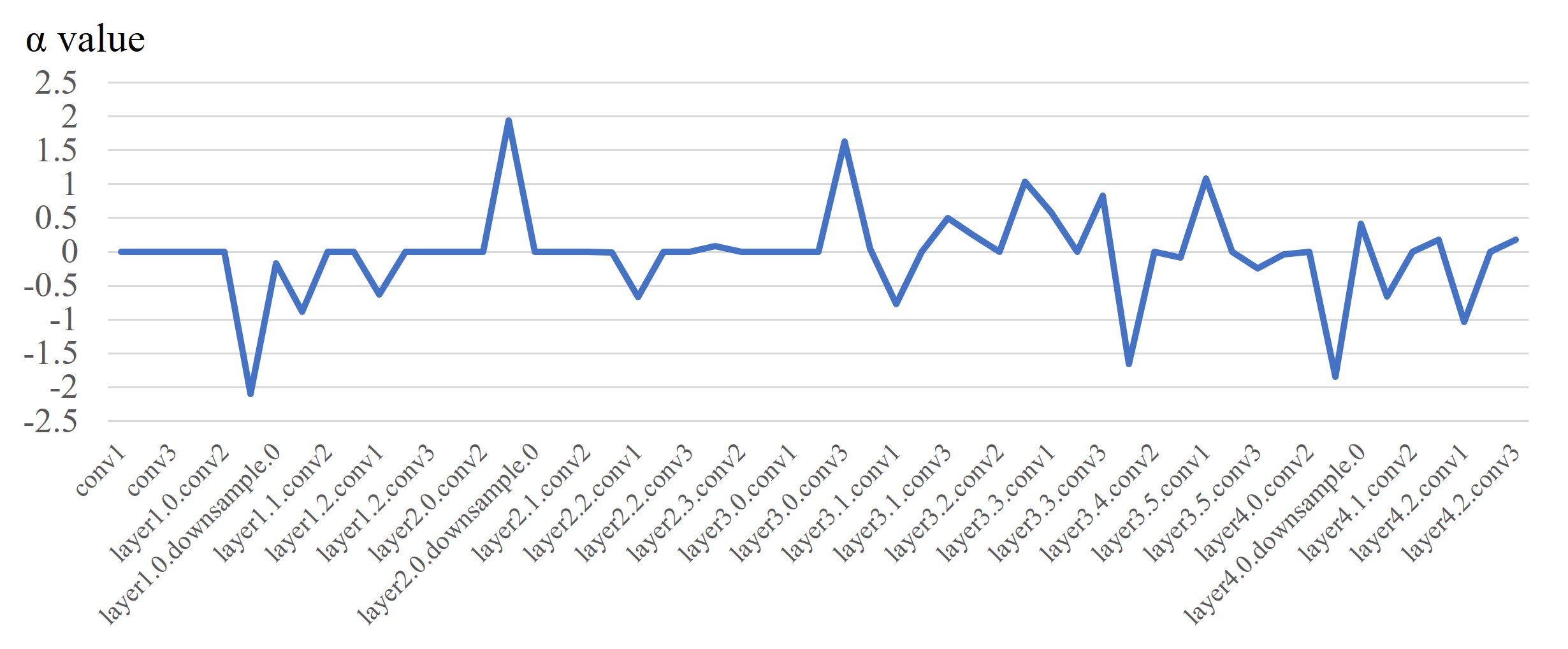}
\end{center}
\vspace{-6mm}
\caption{$\alpha$ changes before and after finetuning.}
\vspace{-6pt}
\label{fig:alpha-change}   
\end{figure} 
\vspace{0.4cm}


\subsection{Ablation Study Implementation details}
In this section, we conduct several ablations to justify the
design choices of our proposed methods.

\noindent \textbf{Compare with other compression methods}
In ~\cref{table:transfer}, we mainly compare our method with random pruning, since it can also be transferred to different models without incurring any additional costs.  
In ~\cref{table:sup_purning_method}, we compare our method with the established pruning technique IMP. Our approach outperforms IMP in terms of both performance and training time. Moreover, our subnetwork exhibits transferability, allowing for faster training across different segment architectures.

\noindent \textbf{OVS capability analysis}
Our core method leverages the benefits of our transferable subnetwork to improve OVS performance and enable enhanced open-set knowledge distillation from CLIP.
As shown in Tab.~\ref{table:cka_ablation}, by adopting our subnetwork, DeeplabV3 can achieve more similar features with CLIP than the baseline, measured by Centered Kernel Alignment (CKA) similarity averaged over layers~\cite{cortes2012algorithms}.

\noindent \textbf{Analysis of freezing layers}
As shown in ~\cref{fig:alpha-change}, compared with early layers, $\alpha$ of deeper layers undergo more changes during fine-tuning, gradually becoming more ``well-trained" as their $\alpha$s decrease~\cite{martin2021implicit}. 

Based on the implicit self-regularization in deep networks~\cite{martin2021implicit}, weight matrices with $\alpha < 2$ are generally considered ``over-trained" and more prone to overfitting. Therefore, in our supplementary Figure 6, we observe certain layers with $\alpha < 2$. Freezing these layers during fine-tuning provides benefits, as it helps prevent overfitting.
We also provide different ratios of freezing layers in ~\cref{table:alpha_ratio_ablation}.
Users can adjust this ratio flexibly according to their own needs.

\subsection{More Qualitative Results}
Building upon the findings in ~\cref{sec:qualitative_results}, we present additional qualitative results in this section, along with a comparison to the ground truth. ~\cref{fig:sup_vis} illustrates a specific case where our model demonstrates robust OVS performance by accurately labeling some parts that are not labeled in the ground truth. This exemplifies the effectiveness of our model in accurately predicting labels even in challenging scenarios where ground truth annotations may be incomplete.

\end{document}